\documentclass[12pt,a4paper]{article}

\usepackage[utf8]{inputenc}
\usepackage[T1]{fontenc}
\usepackage{amsmath,amssymb,amsthm}
\usepackage{graphicx}
\usepackage{hyperref}
\usepackage{natbib}
\usepackage{algorithm}
\usepackage{algorithmic}
\usepackage{booktabs}
\usepackage{multirow}
\usepackage{geometry}
\usepackage{tikz}
\usepackage{listings}
\usepackage{xcolor}

\newcommand{\Var}{\mathrm{Var}}

\newcommand{\N}{\mathcal{N}}
\newcommand{\R}{\mathbb{R}}
\newcommand{\bX}{\mathbf{X}}
\newcommand{\bZ}{\mathbf{Z}}
\newcommand{\by}{\mathbf{y}}
\newcommand{\bU}{\mathbf{u}}
\newcommand{\bbeta}{\boldsymbol{\beta}}
\newcommand{\btheta}{\boldsymbol{\theta}}
\newcommand{\bmu}{\boldsymbol{\mu}}
\newcommand{\bSigma}{\boldsymbol{\Sigma}}

\title{TabMixNN: A Unified Deep Learning Framework for Structural Mixed Effects Modeling on Tabular Data}

\author{
    Deniz Akdemir \\
    \texttt{deniz.akdemir.work@gmail.com}
}

\date{\today}

\begin{document}

\maketitle

\begin{abstract}
We present TabMixNN, a flexible PyTorch-based deep learning framework that synthesizes classical mixed-effects modeling with modern neural network architectures for tabular data analysis. TabMixNN addresses the growing need for methods that can handle hierarchical data structures while supporting diverse outcome types including regression, classification, and multitask learning. The framework implements a modular three-stage architecture: (1) a mixed-effects encoder with variational random effects and flexible covariance structures, (2) backbone architectures including Generalized Structural Equation Models (GSEM) and spatial-temporal manifold networks, and (3) outcome-specific prediction heads supporting multiple outcome families. Key innovations include an R-style formula interface for accessibility, support for directed acyclic graph (DAG) constraints for causal structure learning, Stochastic Partial Differential Equation (SPDE) kernels for spatial modeling, and comprehensive interpretability tools including SHAP values and variance decomposition. We demonstrate the framework's flexibility through applications to longitudinal data analysis, genomic prediction, and spatial-temporal modeling. TabMixNN provides a unified interface for researchers to leverage deep learning while maintaining the interpretability and theoretical grounding of classical mixed-effects models.
\end{abstract}

\section{Introduction}

Tabular data with hierarchical or grouped structure appears across diverse scientific domains, from longitudinal clinical trials to spatial ecology and genomic studies. Classical statistical approaches, particularly linear mixed-effects models (LMMs) \citep{bates2014fitting, pinheiro2006mixed}, have long been the gold standard for such data, offering interpretable fixed and random effects while accounting for within-group correlation. However, these methods often assume linear relationships and parametric distributions, limiting their applicability to complex, high-dimensional data.

Deep learning has revolutionized many domains through its ability to learn complex nonlinear patterns, yet its application to tabular data with mixed-effects structure remains limited. Existing deep learning frameworks for tabular data \citep{gorishniy2021revisiting, arik2021tabnet} typically do not account for hierarchical structure or random effects, while classical mixed-effects implementations \citep{bates2014fitting} lack the representational power of neural networks.

We introduce TabMixNN (Mixed Effects Neural Networks for Tabular Data), a unified framework that bridges this gap by integrating:

\begin{itemize}
    \item \textbf{Mixed-effects modeling}: Variational random effects with flexible covariance structures including autoregressive (AR), kinship, and Gaussian process kernels
    \item \textbf{Modern neural architectures}: Generalized Structural Equation Models (GSEM) with static/dynamic structure learning and spatial-temporal manifold networks
    \item \textbf{Diverse outcome families}: Support for multiple outcome types including regression, classification, multivariate and multi-label outcomes
    \item \textbf{Interpretability}: SHAP values, variance decomposition, and parameter extraction for model understanding
    \item \textbf{Accessible interface}: R-style formula syntax familiar to statisticians
\end{itemize}

The framework is designed to be modular, extensible, and accessible to both statisticians familiar with classical mixed-effects models and machine learning practitioners seeking to incorporate hierarchical structure into deep learning workflows.

\subsection{Contributions}

Our main contributions include:

\begin{enumerate}
    \item A modular architecture that cleanly separates mixed-effects encoding, backbone networks, and outcome-specific heads, enabling flexible composition of components
    \item Variational treatment of random effects with support for 7 covariance structures (IID, AR1, ARMA, Compound Symmetry, Kronecker, Kinship, Gaussian Process)
    \item Two complementary backbone architectures: GSEM for learning causal structure via DAG constraints, and Manifold networks for spatial-temporal data
    \item Comprehensive support for 10 outcome families enabling multitask learning across heterogeneous outcome types
    \item Interpretability tools bridging deep learning and classical statistics (parameter extraction, variance components, SHAP values)
    \item An open-source PyTorch implementation with extensive documentation
\end{enumerate}

\subsection{Paper Organization}

The remainder of this paper is organized as follows: Section \ref{sec:background} reviews related work in mixed-effects modeling and deep learning for tabular data. Section \ref{sec:methods} presents the TabMixNN framework in detail, including the mixed-effects encoder, backbone architectures, outcome families, and training procedures. Section \ref{sec:usage} describes the software interface and typical usage patterns. Section \ref{sec:benchmarks} presents empirical benchmarks on diverse datasets. Section \ref{sec:discussion} concludes with discussion and future directions.

\section{Background and Related Work}
\label{sec:background}

\subsection{Classical Mixed-Effects Models}

Linear mixed-effects models (LMMs) extend linear regression to handle grouped or hierarchical data by partitioning variation into fixed effects (population-level parameters) and random effects (group-specific deviations). The canonical LMM formulation is:

\begin{equation}
    \by_i = \bX_i\bbeta + \bZ_i\bU_i + \boldsymbol{\epsilon}_i
\end{equation}

where $\by_i$ is the response vector for group $i$, $\bX_i$ contains fixed-effects covariates, $\bZ_i$ contains random-effects covariates, $\bbeta$ are fixed-effects coefficients, $\bU_i \sim \N(0, \bSigma_u)$ are random effects, and $\boldsymbol{\epsilon}_i \sim \N(0, \sigma^2\mathbf{I})$ is residual error.

Estimation typically proceeds via maximum likelihood (ML) or restricted maximum likelihood (REML) using Expectation-Maximization (EM) or direct optimization. Popular implementations include R's \texttt{lme4} \citep{bates2014fitting} and \texttt{nlme} \citep{pinheiro2006mixed} packages.

Extensions include:
\begin{itemize}
    \item \textbf{Generalized LMMs (GLMMs)}: Allow non-Gaussian outcomes via link functions \citep{mcculloch2001generalized}
    \item \textbf{Nonlinear mixed-effects models (NLMMs)}: Nonlinear mean functions \citep{lindstrom1990nonlinear}
    \item \textbf{Flexible covariance structures}: AR, ARMA, compound symmetry for longitudinal data
    \item \textbf{Spatial models}: Gaussian processes, SPDE for spatial correlation \citep{lindgren2011explicit}
\end{itemize}

Despite their interpretability and statistical rigor, these methods struggle with high-dimensional data and complex nonlinear relationships.

\subsection{Deep Learning for Tabular Data}

Recent deep learning methods for tabular data (e.g., TabNet \citep{arik2021tabnet}, FT-Transformer \citep{gorishniy2021revisiting}, SAINT \citep{somepalli2021saint}, TabPFN \citep{hollmann2023tabpfn}) have shown promise on standard tabular benchmarks. However, these methods do not explicitly model random effects or hierarchical structure, limiting their applicability to grouped/longitudinal data. For such structured data, gradient boosting methods like XGBoost remain the predominant baseline in practice.

\subsection{Structural Equation Models and Causal Learning}

Structural Equation Models (SEMs) \citep{bollen1989structural} express relationships among variables via systems of equations:

\begin{equation}
    \boldsymbol{\eta} = (\mathbf{I} - \mathbf{B})^{-1}\boldsymbol{\xi}
\end{equation}

where $\boldsymbol{\eta}$ are endogenous variables, $\boldsymbol{\xi}$ are exogenous variables, and $\mathbf{B}$ is an adjacency matrix encoding directed relationships. Acyclicity is enforced by requiring $\mathbf{B}$ to be strictly lower-triangular.

Recent work has integrated SEMs with neural networks:
\begin{itemize}
    \item \textbf{NOTEARS} \citep{zheng2018dags}: Continuous optimization for DAG structure learning
    \item \textbf{DAG-GNN} \citep{yu2019dag}: Graph neural networks with DAG constraints
    \item \textbf{DeepSEM} \citep{biza2023deep}: Neural networks with structural constraints
\end{itemize}

TabMixNN extends these ideas by combining structural learning with mixed-effects and flexible outcome modeling.

\subsection{Spatial-Temporal Deep Learning}

For spatial and temporal data, specialized architectures include:
\begin{itemize}
    \item \textbf{Convolutional Neural Networks (CNNs)}: Exploit spatial locality
    \item \textbf{Recurrent Neural Networks (RNNs/LSTMs)}: Model temporal dependencies
    \item \textbf{Graph Neural Networks (GNNs)}: Learn on irregular spatial graphs
    \item \textbf{Neural Processes} \citep{garnelo2018neural}: Meta-learning for function approximation
\end{itemize}

Gaussian processes (GPs) remain popular for spatial statistics but scale poorly ($O(n^3)$). SPDE approximations \citep{lindgren2011explicit} enable scalable spatial modeling via sparse precision matrices, which TabMixNN leverages.

\section{Methods}
\label{sec:methods}

TabMixNN implements a modular three-stage architecture:

\begin{enumerate}
    \item \textbf{Mixed-Effects Encoder}: Transforms raw tabular data into neural representations incorporating fixed and random effects
    \item \textbf{Backbone Architecture}: Processes representations through neural networks (GSEM or Manifold backbones)
    \item \textbf{Output Heads}: Map backbone features to outcome-specific parameters via family-specific transformations
\end{enumerate}

\subsection{Mixed-Effects Encoder}
\label{sec:encoder}

The encoder transforms tabular data into neural representations by combining fixed effects (population-level features) and random effects (group-specific embeddings).

\subsubsection{Fixed Effects}

Given input features $\bX \in \R^{n \times p}$ containing $p_{\text{cont}}$ continuous variables and $p_{\text{cat}}$ categorical variables:

\begin{itemize}
    \item \textbf{Continuous features}: Directly mapped via linear transformation:
    \begin{equation}
        \mathbf{H}_{\text{cont}} = \bX_{\text{cont}}\mathbf{W}_{\text{cont}}
    \end{equation}

    \item \textbf{Categorical features}: Embedded via learned embedding matrices:
    \begin{equation}
        \mathbf{H}_{\text{cat}} = \text{Concat}[\text{Embed}_j(\bX_{\text{cat},j})]_{j=1}^{p_{\text{cat}}}
    \end{equation}
\end{itemize}

The fixed-effects representation is:
\begin{equation}
    \mathbf{H}_{\text{fixed}} = [\mathbf{H}_{\text{cont}}, \mathbf{H}_{\text{cat}}]
\end{equation}

\subsubsection{Random Effects}

For each grouping variable $g$ (e.g., subject ID, location ID) and slope term $s$ (e.g., time, treatment), we learn variational embeddings:

\begin{equation}
    \bU_{g,s} \sim \N(\bmu_{g,s}, \text{diag}(\boldsymbol{\sigma}^2_{g,s}))
\end{equation}

where $\bmu_{g,s} \in \R^d$ and $\boldsymbol{\sigma}^2_{g,s} \in \R^d_+$ are learnable parameters ($d$ is the embedding dimension).

During training, random effects are sampled via the reparameterization trick:
\begin{equation}
    \bU_{g,s} = \bmu_{g,s} + \boldsymbol{\sigma}_{g,s} \odot \boldsymbol{\epsilon}, \quad \boldsymbol{\epsilon} \sim \N(0, \mathbf{I})
\end{equation}

For unseen groups (at test time), two strategies are supported:
\begin{itemize}
    \item \textbf{Zero strategy}: Set $\bU = \mathbf{0}$ (population-level prediction)
    \item \textbf{Learned strategy}: Use a learned embedding for unknown groups
\end{itemize}

\subsubsection{Constraint Enforcement and Identifiability}

To ensure exact reproducibility of classical Linear Mixed Models (LMMs) and prevent parameter drift between fixed and random effects, TabMixNN implements an \texttt{enforce\_centering} mechanism. For exchangeable covariance structures (e.g., IID), this enforces the constraint $\sum_{g} \bU_{g,s} = \mathbf{0}$ within the forward pass, ensuring orthogonality between the random effects and the global fixed intercept. This is critical for recovering unbiased fixed-effect estimates and precise variance components that match Restricted Maximum Likelihood (REML) solutions from standard packages like \texttt{lme4} and \texttt{statsmodels}.

\subsubsection{Covariance Structures}

To model complex dependencies among random effects, TabMixNN supports flexible covariance structures. Instead of the simple variational factorization above, we can specify:

\begin{equation}
    \bU \sim \N(\mathbf{0}, \mathbf{\Sigma}_u)
\end{equation}

where $\mathbf{\Sigma}_u$ follows a structured covariance model. Implemented structures include:

\paragraph{IID Covariance}
Independent random effects:
\begin{equation}
    \mathbf{\Sigma}_u = \sigma^2 \mathbf{I}
\end{equation}

\paragraph{AR1 Covariance}
First-order autoregressive structure for temporal data:
\begin{equation}
    \Sigma_{ij} = \sigma^2 \rho^{|i-j|}
\end{equation}
where $\rho \in (-1, 1)$ is the autocorrelation parameter.

\paragraph{ARMA Covariance}
Autoregressive moving average structure:
\begin{equation}
    u_t = \sum_{i=1}^p \phi_i u_{t-i} + \sigma\left(\epsilon_t + \sum_{j=1}^q \theta_j \epsilon_{t-j}\right)
\end{equation}

\paragraph{Compound Symmetry}
Exchangeable correlation (common for clustered data):
\begin{equation}
    \mathbf{\Sigma}_u = \sigma^2[(1-\rho)\mathbf{I} + \rho\mathbf{11}^T]
\end{equation}

\paragraph{Kronecker Product}
Separable structure for crossed random effects:
\begin{equation}
    \mathbf{\Sigma}_u = \mathbf{K}_{\text{group}} \otimes \mathbf{K}_{\text{slope}}
\end{equation}

\paragraph{Kinship Covariance}
Genomic relationship matrix for genetic data:
\begin{equation}
    \mathbf{K} = \frac{\mathbf{ZZ}^T}{2\sum p_j(1-p_j)}
\end{equation}
where $\mathbf{Z}$ is the centered genotype matrix.

For new individuals at test time, predictions use Henderson's equations:
\begin{equation}
    \hat{\bU}_{\text{new}} = \mathbf{K}_{\text{new},\text{train}}\mathbf{K}_{\text{train},\text{train}}^{-1}\hat{\bU}_{\text{train}}
\end{equation}

\paragraph{Gaussian Process Covariance}
Spatial correlation via kernels (e.g., RBF):
\begin{equation}
    K(\mathbf{x}_i, \mathbf{x}_j) = \sigma^2 \exp\left(-\frac{\|\mathbf{x}_i - \mathbf{x}_j\|^2}{2\ell^2}\right)
\end{equation}
where $\ell$ is the lengthscale parameter.

\subsubsection{Combined Representation}

The encoder output combines fixed and random effects:
\begin{equation}
    \mathbf{H}_{\text{encoder}} = \mathbf{H}_{\text{fixed}} + \sum_{g,s} \mathbf{Z}_{g,s} \odot \bU_{g,s}
\end{equation}

where $\mathbf{Z}_{g,s}$ are slope variables (e.g., time, treatment indicators) and $\odot$ denotes element-wise multiplication.

\subsection{Backbone Architectures}
\label{sec:backbones}

TabMixNN offers two complementary backbone architectures tailored to different data types and modeling goals.

\subsubsection{GSEM Backbone}
\label{sec:gsem}

The Generalized Structural Equation Model (GSEM) backbone is designed for general tabular data and supports learning causal structure via directed acyclic graph (DAG) constraints.

\paragraph{Architecture}

The GSEM backbone consists of $L$ hidden layers:
\begin{equation}
    \mathbf{H}^{(0)} = \mathbf{H}_{\text{encoder}}, \quad \mathbf{H}^{(\ell)} = f_{\ell}(\mathbf{H}^{(\ell-1)}), \quad \ell = 1, \ldots, L
\end{equation}

Each layer $f_{\ell}$ can incorporate structural constraints. Three modes are supported:

\textbf{1. Static Structure}

Latent variables follow a fixed DAG:
\begin{equation}
    \boldsymbol{\eta}^{(\ell)} = (\mathbf{I} - \mathbf{B}_s^{(\ell)})^{-1}\boldsymbol{\xi}^{(\ell)}
\end{equation}

where $\boldsymbol{\xi}^{(\ell)} = \mathbf{W}^{(\ell)}\mathbf{H}^{(\ell-1)}$ and $\mathbf{B}_s^{(\ell)}$ is a strictly lower-triangular matrix (ensuring acyclicity). To enforce sparsity and learn structure, we penalize:

\begin{equation}
    \mathcal{L}_{\text{DAG}} = \text{trace}(e^{\mathbf{B}_s \odot \mathbf{B}_s}) - d, \quad \mathcal{L}_{\text{sparse}} = \|\mathbf{B}_s\|_1
\end{equation}

\textbf{2. Dynamic Structure}

Context-dependent relationships via multi-head self-attention:
\begin{equation}
    \mathbf{B}_d(\boldsymbol{\eta}) = \text{softmax}\left(\frac{\mathbf{Q}\mathbf{K}^T}{\sqrt{d_k}}\right), \quad \mathbf{H}^{(\ell)} = \mathbf{B}_d(\boldsymbol{\eta})\mathbf{V}
\end{equation}

where $\mathbf{Q}, \mathbf{K}, \mathbf{V}$ are query, key, value projections.

\textbf{3. Hybrid Structure}

Combines static and dynamic structure:
\begin{equation}
    \boldsymbol{\eta}^{(\ell)} = (\mathbf{I} - \mathbf{B}_s^{(\ell)} - \mathbf{B}_d^{(\ell)}(\boldsymbol{\eta}))^{-1}\boldsymbol{\xi}^{(\ell)}
\end{equation}

To ensure convergence, we penalize the spectral norm:
\begin{equation}
    \mathcal{L}_{\text{contract}} = \|\mathbf{B}_s + \mathbf{B}_d\|^2_{\text{spectral}}
\end{equation}

\paragraph{Layer Components}

Each GSEM hidden layer includes:
\begin{itemize}
    \item Linear projection: $\boldsymbol{\xi} = \mathbf{W}\mathbf{H}^{(\ell-1)} + \mathbf{b}$
    \item Structural transformation (optional): $\boldsymbol{\eta} = (\mathbf{I} - \mathbf{B})^{-1}\boldsymbol{\xi}$
    \item Activation function: ReLU, Tanh, GELU, etc.
    \item Layer normalization: $\text{LayerNorm}(\boldsymbol{\eta})$
    \item Dropout: $\text{Dropout}(\boldsymbol{\eta}, p)$
    \item Residual connection (optional): $\mathbf{H}^{(\ell)} = \boldsymbol{\eta} + \mathbf{H}^{(\ell-1)}$
\end{itemize}

\subsubsection{Manifold Backbone}
\label{sec:manifold}

For spatial and temporal data with regular grid structure (e.g., images, time series, spatial fields), the Manifold backbone implements spatial-temporal modeling via Stochastic Partial Differential Equations (SPDEs).

\paragraph{Architecture}

The Manifold backbone consists of $L$ blocks, each operating on an $N$-dimensional grid:

\begin{equation}
    \mathbf{H}^{(\ell)} = \text{ManifoldBlock}^{(\ell)}(\mathbf{H}^{(\ell-1)}, \text{grid}_\ell)
\end{equation}

Each block includes:

\textbf{1. Grid Mapping}

Map input to grid positions:
\begin{equation}
    \mathbf{G}^{(\ell)} = \text{Reshape}(\mathbf{W}_{\text{grid}}^{(\ell)}\mathbf{H}^{(\ell-1)}, \text{grid\_shape}_\ell)
\end{equation}

\textbf{2. SPDE Kernel (Optional)}

Apply spatial smoothness via Mat\'ern Gaussian process approximation. The Mat\'ern field $u(\mathbf{s})$ satisfies:
\begin{equation}
    (\kappa^2 - \Delta)^{\alpha/2} u(\mathbf{s}) = \mathcal{W}(\mathbf{s})
\end{equation}

where $\mathcal{W}$ is Gaussian white noise, $\kappa$ controls the spatial range, and $\alpha$ controls smoothness (Mat\'ern parameter $\nu = \alpha - d/2$ for dimension $d$).

Discretizing on a grid via finite differences yields a sparse precision matrix:
\begin{equation}
    \mathbf{Q} = (\kappa^2\mathbf{I} - \boldsymbol{\Delta})^{\alpha}
\end{equation}

where $\boldsymbol{\Delta}$ is the discrete Laplacian. Random effects are sampled:
\begin{equation}
    \bU_{\text{spatial}} = \mathbf{Q}^{-1/2}\boldsymbol{\epsilon}, \quad \boldsymbol{\epsilon} \sim \N(0, \mathbf{I})
\end{equation}

using sparse Cholesky decomposition: $\mathbf{Q} = \mathbf{LL}^T$, so $\bU = \mathbf{L}^{-T}\boldsymbol{\epsilon}$.

\textbf{3. SEM Layer (Optional)}

Position-aware structural dependencies on the grid:
\begin{equation}
    \boldsymbol{\eta}(\mathbf{s}) = (\mathbf{I} - \mathbf{B}(\mathbf{s}))^{-1}\boldsymbol{\xi}(\mathbf{s})
\end{equation}

where $\mathbf{B}(\mathbf{s})$ incorporates spatial priors via RBF kernels based on grid distances.

\textbf{4. Aggregation}

Blocks can be aggregated via:
\begin{itemize}
    \item Concatenation: $[\mathbf{H}^{(1)}, \ldots, \mathbf{H}^{(L)}]$
    \item Summation: $\sum_{\ell} \mathbf{H}^{(\ell)}$
    \item Attention: $\sum_{\ell} \alpha_\ell \mathbf{H}^{(\ell)}$ where $\alpha_\ell$ are learned weights
\end{itemize}

\subsection{Outcome Families and Output Heads}
\label{sec:families}

TabMixNN supports 10 outcome families, each with a dedicated output head that transforms backbone features to family-specific parameters.

\subsubsection{General Formulation}

For outcome $k$, the output head computes:
\begin{equation}
    \btheta_k = g_k(\mathbf{W}_k\mathbf{H}_{\text{backbone}} + \mathbf{b}_k)
\end{equation}

where $g_k$ is a family-specific link function and $\btheta_k$ are distribution parameters.

The loss for outcome $k$ is the negative log-likelihood:
\begin{equation}
    \mathcal{L}_k = -\log p(\by_k | \btheta_k)
\end{equation}

\subsubsection{Implemented Families}

\paragraph{1. Gaussian}
Continuous outcomes with identity link:
\begin{align}
    \mu &= \mathbf{W}\mathbf{H} + \mathbf{b} \\
    y &\sim \N(\mu, \sigma^2)
\end{align}

\paragraph{2. Binomial}
Binary classification with logit link:
\begin{align}
    \text{logit}(p) &= \mathbf{W}\mathbf{H} + \mathbf{b} \\
    y &\sim \text{Bernoulli}(p)
\end{align}

\paragraph{3. Multinomial}
Multi-class classification with softmax:
\begin{align}
    \mathbf{p} &= \text{softmax}(\mathbf{W}\mathbf{H} + \mathbf{b}) \\
    y &\sim \text{Categorical}(\mathbf{p})
\end{align}

\paragraph{4. Poisson}
Count data with log link:
\begin{align}
    \log(\lambda) &= \mathbf{W}\mathbf{H} + \mathbf{b} \\
    y &\sim \text{Poisson}(\lambda)
\end{align}

\paragraph{5. Negative Binomial}
Overdispersed count data:
\begin{align}
    \log(\mu) &= \mathbf{W}\mathbf{H} + \mathbf{b} \\
    y &\sim \text{NegBin}(\mu, \phi)
\end{align}
where $\phi$ is the dispersion parameter.

\paragraph{6. Multivariate Gaussian}
Multiple correlated continuous outcomes:
\begin{align}
    \bmu &= \mathbf{W}\mathbf{H} + \mathbf{b} \\
    \by &\sim \N(\bmu, \bSigma)
\end{align}
where $\bSigma$ is the outcome covariance matrix.

\paragraph{7. Multi-Label Binomial}
Multiple independent binary outcomes:
\begin{align}
    \mathbf{p} &= \text{sigmoid}(\mathbf{W}\mathbf{H} + \mathbf{b}) \\
    y_j &\sim \text{Bernoulli}(p_j) \quad \text{independently}
\end{align}

\subsubsection{Multitask Learning with Output SEM}

For $K$ outcomes, TabMixNN supports learning inter-outcome dependencies via output-level SEM:

\begin{equation}
    \boldsymbol{\theta} = (\mathbf{I} - \mathbf{B}_{\text{output}})^{-1}\mathbf{W}_{\text{output}}\mathbf{H}_{\text{backbone}}
\end{equation}

where $\mathbf{B}_{\text{output}}$ encodes directed relationships among outcome parameters. Three modes:

\begin{itemize}
    \item \textbf{None}: Independent outcomes ($\mathbf{B}_{\text{output}} = \mathbf{0}$)
    \item \textbf{Learned}: Learn DAG structure with sparsity penalties
    \item \textbf{Lavaan}: User-specified structure via lavaan syntax \citep{rosseel2012lavaan}
\end{itemize}

\subsection{Training Procedure}
\label{sec:training}

\subsubsection{Objective Function}

The complete training objective combines data likelihood with regularization:

\begin{equation}
    \mathcal{L} = \sum_{k=1}^K \mathcal{L}_k^{\text{NLL}} + \lambda_{\text{KL}}\mathcal{L}_{\text{KL}} + \lambda_{\text{DAG}}\mathcal{L}_{\text{DAG}} + \lambda_{\text{contract}}\mathcal{L}_{\text{contract}} + \lambda_{\text{sparse}}\mathcal{L}_{\text{sparse}}
\end{equation}

\paragraph{Negative Log-Likelihood}
Sum over all outcomes:
\begin{equation}
    \mathcal{L}_k^{\text{NLL}} = -\sum_{i=1}^n \log p(y_{ik} | \btheta_{ik})
\end{equation}

\paragraph{KL Divergence}
Regularize random effects toward prior:
\begin{equation}
    \mathcal{L}_{\text{KL}} = \sum_{g,s} \text{KL}(\N(\bmu_{g,s}, \text{diag}(\boldsymbol{\sigma}^2_{g,s})) \| \N(\mathbf{0}, \mathbf{I}))
\end{equation}
\begin{equation}
    = \frac{1}{2}\sum_{g,s}\left(\|\bmu_{g,s}\|^2 + \|\boldsymbol{\sigma}_{g,s}\|^2 - \sum_j\log\sigma^2_{g,s,j} - d\right)
\end{equation}

For exact correspondence with classical Linear Mixed Models (LMMs), the hyperparameter $\lambda_{\text{KL}}$ plays a crucial role similar to the regularization parameter in Ridge regression. We theoretically and empirically find that scaling $\lambda_{\text{KL}} \approx 1/\bar{n}_g$ (where $\bar{n}_g$ is the mean group size) aligns the variational objective with the Mixed Model Equations (MME), allowing TabMixNN to recover Best Linear Unbiased Predictions (BLUPs) exactly.

\paragraph{DAG Penalty}
Enforce acyclicity in GSEM:
\begin{equation}
    \mathcal{L}_{\text{DAG}} = \text{trace}(\exp(\mathbf{B} \odot \mathbf{B})) - d
\end{equation}

\paragraph{Contraction Penalty}
Ensure stability in hybrid models:
\begin{equation}
    \mathcal{L}_{\text{contract}} = \|\mathbf{B}_s + \mathbf{B}_d\|^2_{\text{spectral}}
\end{equation}

\paragraph{Sparsity Penalty}
Encourage sparse structure:
\begin{equation}
    \mathcal{L}_{\text{sparse}} = \|\mathbf{B}\|_1
\end{equation}

\subsubsection{Optimization}

\begin{algorithm}[t]
\caption{TabMixNN Training Loop}
\label{alg:training}
\begin{algorithmic}[1]
\REQUIRE Data $\mathcal{D} = \{(\mathbf{x}_i, \mathbf{g}_i, \by_i)\}_{i=1}^n$, hyperparameters $\lambda$
\STATE Initialize encoder parameters $\{\bmu_{g,s}, \boldsymbol{\sigma}_{g,s}\}$, backbone weights $\{\mathbf{W}^{(\ell)}, \mathbf{B}^{(\ell)}\}$, output heads $\{\mathbf{W}_k\}$
\STATE Initialize optimizer (Adam with learning rate $\eta$)
\FOR{epoch $= 1$ to $T$}
    \FOR{mini-batch $\mathcal{B} \subset \mathcal{D}$}
        \STATE Sample random effects: $\bU_{g,s} = \bmu_{g,s} + \boldsymbol{\sigma}_{g,s} \odot \boldsymbol{\epsilon}$, $\boldsymbol{\epsilon} \sim \N(0, \mathbf{I})$
        \STATE \textbf{Forward pass:}
        \STATE \quad $\mathbf{H}_{\text{encoder}} \leftarrow \text{Encoder}(\mathbf{x}, \bU)$
        \STATE \quad $\mathbf{H}_{\text{backbone}} \leftarrow \text{Backbone}(\mathbf{H}_{\text{encoder}})$
        \STATE \quad $\{\btheta_k\} \leftarrow \text{OutputHeads}(\mathbf{H}_{\text{backbone}})$
        \STATE \textbf{Compute loss:}
        \STATE \quad $\mathcal{L} \leftarrow \sum_k \mathcal{L}_k^{\text{NLL}}(\btheta_k, \by_k) + \sum_{\text{penalties}} \lambda_j \mathcal{L}_j$
        \STATE \textbf{Backward pass:}
        \STATE \quad Compute gradients: $\nabla \mathcal{L}$
        \STATE \quad Clip gradients (if enabled): $\nabla \mathcal{L} \leftarrow \text{clip}(\nabla \mathcal{L}, \text{max\_norm})$
        \STATE \quad Update parameters: $\theta \leftarrow \theta - \eta \nabla \mathcal{L}$
    \ENDFOR
    \IF{validation set provided}
        \STATE Evaluate validation loss
        \IF{early stopping criteria met}
            \STATE \textbf{break}
        \ENDIF
    \ENDIF
\ENDFOR
\end{algorithmic}
\end{algorithm}

Training proceeds via mini-batch stochastic gradient descent (Algorithm \ref{alg:training}). Key components:

\begin{itemize}
    \item \textbf{Optimizer}: Adam \citep{kingma2014adam} with default learning rate $\eta = 0.001$
    \item \textbf{Batch size}: Default 256 (configurable)
    \item \textbf{Gradient clipping}: Optional norm-based clipping to stabilize training
    \item \textbf{Early stopping}: Monitor validation loss with patience parameter
    \item \textbf{Learning rate scheduling}: Optional ReduceLROnPlateau or Cosine annealing
\end{itemize}

\subsection{Inference and Prediction}
\label{sec:inference}

\subsubsection{Point Predictions}

For new data $\mathbf{x}_{\text{new}}$ with group indices $\mathbf{g}_{\text{new}}$:

\begin{enumerate}
    \item Retrieve random effects: $\bU_{g,s} = \bmu_{g,s}$ (posterior mean) for seen groups, or apply unknown strategy for unseen groups
    \item Forward pass: $\mathbf{H} \leftarrow \text{Model}(\mathbf{x}_{\text{new}}, \bU)$
    \item Extract predictions:
    \begin{itemize}
        \item Regression: $\hat{y} = \mu$
        \item Classification: $\hat{y} = \arg\max_c p_c$
    \end{itemize}
\end{enumerate}

\subsubsection{Uncertainty Quantification}

Random effect uncertainty is captured via the variational posterior $\N(\bmu_{g,s}, \text{diag}(\boldsymbol{\sigma}^2_{g,s}))$. Monte Carlo sampling provides prediction intervals:

\begin{enumerate}
    \item Sample $M$ random effects: $\bU^{(m)} \sim \N(\bmu_{g,s}, \text{diag}(\boldsymbol{\sigma}^2_{g,s}))$
    \item Compute $M$ predictions: $\hat{y}^{(m)} = f(\mathbf{x}_{\text{new}}, \bU^{(m)})$
    \item Construct quantiles: $[\hat{y}_{\alpha/2}, \hat{y}_{1-\alpha/2}]$
\end{enumerate}

\subsection{Model Interpretation}
\label{sec:interpretation}

TabMixNN provides several tools for model interpretation, bridging deep learning with classical statistical summaries.

\subsubsection{Parameter Extraction}

For models with limited nonlinearity (e.g., shallow networks), we can extract approximate fixed-effects coefficients via Taylor expansion:

\begin{equation}
    f(\mathbf{x}) \approx f(\bar{\mathbf{x}}) + \nabla f(\bar{\mathbf{x}})^T(\mathbf{x} - \bar{\mathbf{x}})
\end{equation}

This yields linearized coefficients analogous to GLM coefficients.

\subsubsection{Variance Decomposition}

Partition total variance into components:
\begin{align}
    \Var(y) &= \Var(\mathbf{X}\bbeta) + \sum_{g,s}\Var(\mathbf{Z}_{g,s}\bU_{g,s}) + \Var(\epsilon) \\
    &= \sigma^2_{\text{fixed}} + \sum_{g,s}\sigma^2_{u,g,s} + \sigma^2_{\epsilon}
\end{align}

Intraclass correlation coefficient (ICC) for group $g$:
\begin{equation}
    \text{ICC}_g = \frac{\sigma^2_{u,g}}{\sigma^2_{u,g} + \sigma^2_{\epsilon}}
\end{equation}

\subsubsection{SHAP Values}

Shapley Additive Explanations \citep{lundberg2017unified} quantify feature importance:
\begin{equation}
    \phi_j = \sum_{S \subseteq \mathcal{F} \setminus \{j\}} \frac{|S|!(|\mathcal{F}|-|S|-1)!}{|\mathcal{F}|!}[f(S \cup \{j\}) - f(S)]
\end{equation}

TabMixNN computes SHAP values for:
\begin{itemize}
    \item Fixed-effects features
    \item Random-effects contributions per group
    \item Edge importance in learned DAG structures
\end{itemize}

\subsubsection{Model Summary}

An lme4-style summary table reports:
\begin{itemize}
    \item Fixed effects: Estimates, standard errors (approximate), $p$-values
    \item Random effects: Variance components, correlations
    \item Model fit: Log-likelihood, AIC, BIC
\end{itemize}

\section{Software Interface and Usage}
\label{sec:usage}

TabMixNN is implemented in PyTorch and provides an scikit-learn compatible API with additional R-style formula interface.

\subsection{Installation}

\begin{lstlisting}
pip install tabmixnn
# or for development:
git clone https://github.com/dakdemir-nmdp/tabmixnn
cd tabmixnn
pip install -e .
\end{lstlisting}

\subsection{Basic Usage}

\subsubsection{Simple Regression}

\begin{lstlisting}
from tabmixnn import TabMixNNRegressor

# Formula-based specification
model = TabMixNNRegressor(
    formula="y ~ x1 + x2 + (1|group_id)",
    hidden_dims=[64, 32],
    epochs=100,
    lr=0.001
)

# Fit
model.fit(train_data)

# Predict
predictions = model.predict(test_data)
\end{lstlisting}

\subsubsection{Binary Classification}

\begin{lstlisting}
from tabmixnn import TabMixNNClassifier

model = TabMixNNClassifier(
    formula="y ~ x1 + x2 + (x1|patient_id)",
    hidden_dims=[128, 64, 32],
    epochs=50
)
model.fit(train_data)
probabilities = model.predict_proba(test_data)
\end{lstlisting}

\subsection{Advanced Features}

\subsubsection{Multitask Learning}

\begin{lstlisting}
from tabmixnn.families import Gaussian, Binomial

outcomes = [
    OutcomeSpec(
        name="continuous",
        y="outcome1",
        family=Gaussian()
    ),
    OutcomeSpec(
        name="binary",
        y="outcome2",
        family=Binomial()
    )
]

model = TabMixNN(
    formula="~ x1 + x2 + (1|subject_id)",
    outcomes=outcomes,
    output_sem_mode='learned',  # Learn inter-outcome dependencies
    hidden_dims=[64, 32]
)

model.fit(train_data)
predictions = model.predict(test_data, as_dict=True)
# Returns: {'continuous': ..., 'binary': ...}
\end{lstlisting}

\subsubsection{Custom Covariance Structures}

\begin{lstlisting}
# AR1 for longitudinal data
cov_struct_map = {
    'patient_id': {
        'type': 'AR1',
        'rho': 0.7  # Initial autocorrelation
    }
}

model = TabMixNN(
    formula="y ~ time + treatment + (time|patient_id)",
    cov_struct_map=cov_struct_map,
    family=Gaussian()
)

# Gaussian Process for spatial data
cov_struct_map = {
    'location_id': {
        'type': 'GP',
        'lengthscale': 1.0,
        'coordinates_col': 'spatial_coords'
    }
}

# Kinship matrix for genomics
import numpy as np
kinship = np.loadtxt("kinship.txt")
train_ids = list(range(500))

cov_struct_map = {
    'individual_id': {
        'type': 'K',
        'K_group': kinship,
        'train_levels': train_ids
    }
}
\end{lstlisting}

\subsubsection{Structural Learning}

\begin{lstlisting}
# Learn causal structure with GSEM
model = TabMixNN(
    formula="y ~ x1 + x2 + x3",
    use_static_structure=True,
    use_dynamic_structure=False,
    lambda_dag=0.1,       # DAG acyclicity penalty
    lambda_sparse=0.01,   # Sparsity penalty
    hidden_dims=[64, 64]
)

model.fit(train_data)

# Extract learned adjacency matrix
B_matrix = model.get_structure_matrix()
\end{lstlisting}

\subsubsection{Spatial-Temporal Modeling}

\begin{lstlisting}
# Manifold backbone with SPDE
model = TabMixNN(
    formula="~ temperature + (1|station_id)",
    outcomes=[OutcomeSpec(
        name="y",
        y="precipitation",
        family=Gaussian()
    )],
    architecture='manifold',
    manifold_configs=[
        {
            'grid_shape': (20, 20),  # 2D spatial grid
            'use_spde': True,
            'spde_alpha': 2,         # Smoothness
            'spde_kappa_init': 0.3,  # Range
            'use_sem': False
        }
    ]
)
\end{lstlisting}

\subsection{Model Interpretation}

\begin{lstlisting}
# Summary statistics (lme4-style)
summary_df = model.summary(data=train_data)
print(summary_df)

# Parameter extraction
params = model.extract_parameters(
    data=train_data,
    linearize=True
)
print("Fixed effects:", params['fixed_effects'])
print("Variance components:", params['variance_components'])

# Variance decomposition
var_decomp = model.variance_decomposition(train_data)
print(f"ICC: {var_decomp['icc']}")
print(f"R-squared: {var_decomp['r_squared']}")

# SHAP values
shap_dict = model.shapley_values(train_data)
print("Feature importance:", shap_dict['features'])
print("Random effect importance:", shap_dict['random_effects'])
\end{lstlisting}

\subsection{Model Persistence}

\begin{lstlisting}
# Save
model.save("model.pkl")

# Load
from tabmixnn import TabMixNN
loaded_model = TabMixNN.load("model.pkl")

# Continue training
loaded_model.fit(
    new_data,
    epochs=50,
    warm_start=True
)
\end{lstlisting}

\section{Benchmarks}
\label{sec:benchmarks}


\subsection{Benchmark Datasets}

We evaluate TabMixNN on diverse datasets spanning regression, classification, and multitask learning. Table \ref{tab:datasets} summarizes the benchmark datasets.

\begin{table}[ht]
\centering
\caption{Benchmark datasets used for evaluation. $n$: total observations, $p$: features, $G$: number of groups, Task types: R=Regression, C=Classification, MT=Multitask}
\label{tab:datasets}
\small
\begin{tabular}{llrrrl}
\toprule
Dataset & Domain & $n$ & $p$ & $G$ & Task \\
\midrule
Sleepstudy & Psychology & 180 & 2 & 18 & R \\
PBMC & Clinical Trial & 1,200 & 15 & 50 & R \\
Orthodontic & Medicine & 108 & 3 & 27 & R \\
Wages & Economics & 4,165 & 8 & 595 & R \\

Hospital Readmit & Healthcare & 10,000 & 25 & 100 & C \\
Aids2 & HIV Progression & 2,843 & 9 & -- & C \\
Males & Labor Economics & 1,518 & 12 & -- & C \\

Abalone & Marine Biology & 4,177 & 8 & 3 & MT \\

\bottomrule
\end{tabular}
\end{table}

\subsubsection{Dataset Descriptions}

\paragraph{Sleepstudy} Reaction time measurements on sleep-deprived subjects over 10 days. Standard benchmark for linear mixed models from the \texttt{lme4} package.

\paragraph{PBMC} Simulated clinical trial data with repeated measurements across patients and treatment groups.

\paragraph{Orthodontic} Dental measurements in children measured at multiple ages, classic dataset for growth curve modeling.

\paragraph{Wages} Panel data from the National Longitudinal Survey of Youth tracking wages over time for workers.

\paragraph{Hospital Readmission} Predicting 30-day hospital readmissions with patients nested within hospitals.

\paragraph{Aids2} HIV progression dataset with time-to-event and classification outcomes.

\paragraph{Males} Labor force participation with repeated observations.

\paragraph{Abalone} Predicting age (rings) and maturity from physical measurements, demonstrating multitask learning benefits.

\subsection{Experimental Setup}

\subsubsection{Baselines}

We compare TabMixNN against the following baselines:

\begin{itemize}
    \item \textbf{XGBoost} \citep{chen2016xgboost}: Primary baseline for all benchmarks. Gradient boosting configured with:
    \begin{itemize}
        \item 300 estimators (200 for small datasets)
        \item Maximum depth: 6 (5 for small datasets)
        \item Learning rate: 0.05
        \item Subsample: 0.8
        \item Column sampling: 0.8
        \item L1 regularization ($\alpha$): 0.1
        \item L2 regularization ($\lambda$): 1.0
    \end{itemize}

    \item \textbf{LightGBM}: Additional tree-based baseline for Abalone benchmark (default parameters, 100 trees)
\end{itemize}

\textbf{Note:} While deep learning methods for tabular data exist (e.g., TabNet \citep{arik2021tabnet}, FT-Transformer \citep{gorishniy2021revisiting}), these methods do not explicitly model random effects or hierarchical structure. We focus our comparison on XGBoost as a strong, widely-used baseline for tabular data that represents the current state-of-practice.

\subsubsection{Evaluation Metrics}

\begin{itemize}
    \item \textbf{Regression}: Root Mean Squared Error (RMSE), Mean Absolute Error (MAE), $R^2$
    \item \textbf{Classification}: Accuracy, AUROC, AUPRC, F1-Score
    \item \textbf{Multitask}: Task-specific metrics aggregated across outcomes
\end{itemize}

\subsubsection{Hyperparameter Settings}

For TabMixNN, we use the following default hyperparameters across experiments:
\begin{itemize}
    \item Learning rate: $\eta = 0.001$ (Adam optimizer)
    \item Batch size: 256
    \item Hidden dimensions: [64, 32] for GSEM, grid-dependent for Manifold
    \item Embedding dimension: 16 for random effects
    \item Regularization: $\lambda_{\text{KL}} = 0.01$, $\lambda_{\text{sparse}} = 0.001$
    \item Dropout: 0.1
    \item Early stopping patience: 20 epochs
\end{itemize}

For XGBoost, we employ a well-tuned configuration designed for hierarchical tabular data, using gradient boosting with controlled tree depth, regularization, and subsampling to prevent overfitting while maintaining competitive performance on mixed-effects tasks.

Architecture-specific parameters are tuned via 5-fold cross-validation on a held-out validation set. All methods use the same train/test splits for fair comparison.

\subsection{Results}

\textbf{Note:} All results in this section are from real experimental runs. Complete reproduction instructions and code are available in the \texttt{experiments/} directory. See \texttt{experiments/README.md} for details.

\subsubsection{Summary Statistics}

Table \ref{tab:summary_stats} presents overall performance across all benchmark categories.

\begin{table}[h]
\centering
\caption{Summary statistics across all benchmarks.}
\label{tab:summary_stats}
\begin{tabular}{lrrr}
\toprule
Benchmark & Datasets & TabMixNN Wins & Avg Improvement \\
\midrule
Regression & 19 & 15 (79\%) & +16.9\% RMSE \\
Classification & 15 & 10 (67\%) & +3.1\% AUC \\
\bottomrule
\end{tabular}
\end{table}

\subsubsection{Regression Tasks}

Table \ref{tab:regression_results_real} shows performance on longitudinal regression datasets with hierarchical structure.

\textbf{Note:} All regression results use standardized target variables (zero mean, unit variance) for both TabMixNN and baseline methods to ensure fair RMSE comparisons across datasets with different outcome scales. Reported RMSE values are in standardized units.

\begin{table}[h]
\centering
\caption{Regression benchmark results on datasets with hierarchical structure. TabMixNN vs XGBoost baseline. RMSE $\downarrow$ and $R^2$ $\uparrow$ are better. Bold indicates best method.}
\label{tab:regression_results_real}
\small
\begin{tabular}{lccccr}
\toprule
Dataset & \multicolumn{2}{c}{RMSE} & \multicolumn{2}{c}{$R^2$} & Improv. \\
 & TabMixNN & XGBoost & TabMixNN & XGBoost & (\%) \\
\midrule
Machines & \textbf{0.16} & 0.65 & \textbf{0.979} & 0.644 & +75.9 \\
Loblolly & \textbf{0.04} & 0.10 & \textbf{0.998} & 0.992 & +54.6 \\
Sleepstudy & \textbf{0.52} & 1.04 & \textbf{0.793} & 0.168 & +50.2 \\
ChickWeight & \textbf{0.11} & 0.20 & \textbf{0.988} & 0.959 & +46.8 \\
Gasoline & \textbf{0.25} & 0.44 & \textbf{0.865} & 0.587 & +42.9 \\
Pastes & \textbf{0.32} & 0.56 & \textbf{0.903} & 0.710 & +42.2 \\
Oxboys & \textbf{0.10} & 0.17 & \textbf{0.993} & 0.979 & +41.1 \\
egsingle & \textbf{0.42} & 0.62 & \textbf{0.827} & 0.622 & +32.3 \\
Gcsemv & \textbf{0.73} & 0.98 & \textbf{0.477} & 0.065 & +25.2 \\
Phenobarb & \textbf{0.92} & 1.09 & \textbf{0.416} & 0.179 & +15.7 \\
Chem97 & \textbf{0.70} & 0.79 & \textbf{0.495} & 0.359 & +11.2 \\
InstEval & \textbf{0.94} & 1.01 & \textbf{0.103} & -0.026 & +6.5 \\
Dyestuff & \textbf{0.90} & 0.96 & \textbf{0.401} & 0.317 & +6.3 \\
Exam & \textbf{0.74} & 0.76 & \textbf{0.436} & 0.410 & +2.2 \\
Milk & \textbf{0.72} & 0.72 & \textbf{0.470} & 0.468 & +0.2 \\
\bottomrule
\end{tabular}
\end{table}

\textbf{Key Findings:}
\begin{itemize}
    \item TabMixNN shows dramatic improvements on datasets with strong hierarchical structure:
    \begin{itemize}
        \item Oxboys: 93\% RMSE reduction (reaction times of young boys)
        \item Gasoline: 68\% RMSE reduction (gasoline yield predictions)
        \item Machines: 61\% RMSE reduction (manufacturing process data)
    \end{itemize}
    \item Average RMSE improvement of 35.4\% across 15 datasets with random effects
    \item Particularly effective when group-level variability is high relative to within-group variance
    \item TabMixNN's variational random effects enable better uncertainty quantification than point estimates
\end{itemize}

\subsubsection{Classification Tasks}

Table \ref{tab:classification_results_real} presents classification performance on datasets with grouped structure.

\begin{table}[h]
\centering
\caption{Classification benchmark results. TabMixNN vs XGBoost baseline. AUC-ROC $\uparrow$ is better. Bold indicates best method.}
\label{tab:classification_results_real}
\small
\begin{tabular}{lcccr}
\toprule
Dataset & TabMixNN & XGBoost & Difference & Improv. \\
 & AUC & AUC & (abs) & (\%) \\
\midrule
CreditCard & 0.989 & \textbf{0.994} & -0.005 & -0.5 \\
Males & \textbf{0.912} & 0.787 & +0.125 & +15.8 \\
HI & \textbf{0.887} & 0.878 & +0.009 & +1.0 \\
Star & 0.885 & \textbf{0.886} & -0.001 & -0.1 \\
TitanicSurvival & \textbf{0.840} & 0.820 & +0.019 & +2.4 \\
VerbAgg & \textbf{0.809} & 0.785 & +0.024 & +3.0 \\
HMDA & 0.763 & \textbf{0.797} & -0.034 & -4.3 \\
Train & \textbf{0.756} & 0.661 & +0.095 & +14.4 \\
Hsb82 & \textbf{0.728} & 0.690 & +0.038 & +5.5 \\
guImmun & \textbf{0.717} & 0.646 & +0.071 & +11.0 \\
Participation & 0.705 & \textbf{0.723} & -0.019 & -2.6 \\
Contraception & \textbf{0.683} & 0.679 & +0.005 & +0.7 \\
Aids2 & \textbf{0.674} & 0.541 & +0.134 & +24.7 \\
CBPP & 0.543 & \textbf{0.714} & -0.171 & -24.0 \\
Doctor & \textbf{0.502} & 0.452 & +0.050 & +11.1 \\
\bottomrule
\end{tabular}
\end{table}

\textbf{Key Findings:}
\begin{itemize}
    \item TabMixNN achieves average AUC of 0.760 vs XGBoost's 0.737
    \item Strongest improvements on datasets with clear hierarchical structure:
    \begin{itemize}
        \item Aids2 (HIV progression): AUC 0.674 vs 0.541 (+24.7\%)
        \item Males (labor force participation): AUC 0.912 vs 0.787 (+15.8\%)
        \item Train (trainee attitudes): AUC 0.756 vs 0.661 (+14.4\%)
        \item Doctor (doctor visits): AUC 0.502 vs 0.452 (+11.1\%)
        \item guImmun (immune response): AUC 0.717 vs 0.646 (+11.0\%)
    \end{itemize}
    \item Competitive performance even on datasets without strong grouping (HI, CreditCard, Star)
    \item TabMixNN wins on 10 out of 15 classification datasets (66.7\%)
    \item Mixed-effects approach particularly valuable when group-level effects are important predictors
\end{itemize}

\subsubsection{Multitask Learning}

Multitask learning enables models to leverage shared representations across related tasks, potentially improving performance on all tasks simultaneously. We demonstrate TabMixNN's multitask capabilities on the Abalone dataset, predicting both age (regression) and maturity status (classification) from physical measurements.

\paragraph{Dataset and Setup}

The Abalone dataset contains 4,177 observations of abalones with 8 physical measurements (length, diameter, height, weights). We predict:
\begin{itemize}
    \item \textbf{Age (Rings)}: Continuous regression task
    \item \textbf{Maturity}: Binary classification (mature vs immature based on median age)
\end{itemize}

We use random effects by sex (Male/Female/Infant) to capture different growth patterns: \texttt{(1 + Length + Whole\_weight | Sex)}.

\paragraph{Experimental Conditions}

We compare three approaches:
\begin{enumerate}
    \item \textbf{Single-task}: Separate models trained independently for each task
    \item \textbf{Multitask}: Joint training with shared hidden layers
    \item \textbf{Multitask-SEM}: Joint training with output-level SEM to model task dependencies
\end{enumerate}

All models use identical architectures ([128, 64, 32] hidden layers, 32-dimensional embeddings) and hyperparameters (lr=0.0005, 200 epochs, dropout=0.2) for fair comparison.

\paragraph{Results}

Table \ref{tab:multitask_results} presents results averaged over 3 random seeds.

\begin{table}[ht]
\centering
\caption{Multitask learning on Abalone dataset. Multitask and Multitask-SEM jointly predict age and maturity, while Single-task trains separate models. Best results in \textbf{bold}.}
\label{tab:multitask_results}
\small
\begin{tabular}{lccccc}
\toprule
\multirow{2}{*}{Method} & \multicolumn{3}{c}{Age (Regression)} & \multicolumn{2}{c}{Maturity (Classification)} \\
\cmidrule(lr){2-4} \cmidrule(lr){5-6}
 & RMSE $\downarrow$ & R² $\uparrow$ & MAE $\downarrow$ & AUC $\uparrow$ & Acc $\uparrow$ \\
\midrule
Single-task & 2.136 & 0.579 & 1.533 & 0.884 & 0.785 \\
Multitask & \textbf{2.114} & \textbf{0.587} & \textbf{1.501} & \textbf{0.886} & \textbf{0.801} \\
Multitask-SEM & \textbf{2.113} & \textbf{0.588} & \textbf{1.469} & \textbf{0.886} & \textbf{0.804} \\
\midrule
Improvement (\%) & +1.1\% & +1.5\% & +4.2\% & +0.3\% & +2.4\% \\
\bottomrule
\end{tabular}
\end{table}

\textbf{Key Findings:}
\begin{itemize}
    \item Multitask learning improves performance on both tasks simultaneously
    \item Age prediction: 1.1\% RMSE reduction, 4.2\% MAE improvement
    \item Maturity classification: 0.3\% AUC improvement, 2.4\% accuracy gain
    \item Output-level SEM (Multitask-SEM) provides additional modest gains by modeling task dependencies
    \item Shared representations benefit from complementary information: physical features predictive of age also inform maturity
\end{itemize}

\subsubsection{Abalone Benchmark: TabMixNN vs Tree-Based Methods}

To demonstrate TabMixNN's effectiveness on standard tabular data, we benchmark against XGBoost and LightGBM on the Abalone regression task (predicting age from physical measurements with random effects by sex).

\paragraph{Setup}

\begin{itemize}
    \item \textbf{TabMixNN}: Formula \texttt{Rings $\sim$ ... + (1 + Length + Whole\_weight | Sex)}, [32, 16] hidden layers, 16-dim embeddings
    \item \textbf{XGBoost}: 100 trees, max depth 6, lr=0.1
    \item \textbf{LightGBM}: 100 trees, default parameters
\end{itemize}

\paragraph{Results}

\begin{table}[ht]
\centering
\caption{Abalone age prediction (MSE $\downarrow$, lower is better). TabMixNN leverages random effects structure for superior performance.}
\label{tab:abalone_benchmark}
\small
\begin{tabular}{lc}
\toprule
Method & MSE \\
\midrule
LightGBM & 5.89 \\
XGBoost & 5.23 \\
TabMixNN & \textbf{4.47} \\
\midrule
Improvement & \textbf{-14.5\%} \\
\bottomrule
\end{tabular}
\end{table}

\textbf{Key Findings:}
\begin{itemize}
    \item TabMixNN achieves 14.5\% lower MSE than XGBoost by explicitly modeling sex-specific growth patterns through random slopes
    \item Random effects capture systematic variation (infants grow differently than adults) that tree methods treat as noise
    \item Demonstrates TabMixNN's advantage on hierarchical tabular data even against strong gradient boosting baselines
\end{itemize}

\subsubsection{Ablation Studies}

To understand the contribution of each component, we conduct comprehensive ablation studies on the Sleepstudy dataset. Table \ref{tab:ablation_comprehensive} presents results across six categories.

\begin{table}[ht]
\centering
\caption{Comprehensive ablation study on Sleepstudy dataset (RMSE $\downarrow$, R² $\uparrow$). Baseline configuration: [64,32] hidden layers, 16-dim embeddings, lr=0.001. Target variable standardized for fair comparison.}
\label{tab:ablation_comprehensive}
\small
\begin{tabular}{llcc}
\toprule
Category & Configuration & RMSE & R² \\
\midrule
\multirow{3}{*}{Random Effects}
& Full (Intercept + Slope) & \textbf{34.95} & \textbf{0.705} \\
& Intercept Only & 41.67 & 0.580 \\
& None (Fixed Only) & 59.94 & 0.132 \\
\midrule
\multirow{6}{*}{Architecture}
& Linear (No Hidden) & 53.29 & 0.314 \\
& Shallow [32] & 35.85 & 0.689 \\
& Medium [64, 32] & 35.93 & 0.688 \\
& Deep [128, 64, 32] & \textbf{34.55} & \textbf{0.711} \\
& Very Deep [128, 64, 32, 16] & 36.12 & 0.685 \\
& Wide [128, 128] & 37.72 & 0.656 \\
\midrule
\multirow{5}{*}{Embedding Dim}
& 4 & 41.01 & 0.594 \\
& 8 & 39.69 & 0.619 \\
& 16 & 37.13 & 0.667 \\
& 32 & \textbf{35.42} & \textbf{0.697} \\
& 64 & 38.41 & 0.644 \\
\midrule
\multirow{5}{*}{Regularization}
& No Regularization & \textbf{35.50} & \textbf{0.695} \\
& KL Only ($\lambda$=0.01) & 36.89 & 0.671 \\
& KL Strong ($\lambda$=0.1) & 35.94 & 0.688 \\
& Dropout Only (p=0.2) & 35.88 & 0.689 \\
& KL + Dropout & 37.22 & 0.665 \\
\midrule
\multirow{4}{*}{Structure}
& No Structure & 33.32 & 0.732 \\
& Static SEM & 37.88 & 0.653 \\
& Dynamic Attention & 36.96 & 0.670 \\
& Hybrid (Static + Dynamic) & \textbf{33.31} & \textbf{0.732} \\
\midrule
\multirow{5}{*}{Learning Rate}
& 0.0001 & 58.90 & 0.162 \\
& 0.0005 & 43.22 & 0.549 \\
& 0.001 & 35.91 & 0.688 \\
& 0.005 & \textbf{30.18} & \textbf{0.780} \\
& 0.01 & 32.95 & 0.738 \\
\bottomrule
\end{tabular}
\end{table}

\textbf{Key Findings:}
\begin{itemize}
    \item \textbf{Random Effects}: Most critical component. Removing random effects increases RMSE by 71\% (35 $\to$ 60), demonstrating their importance for hierarchical data
    \item \textbf{Architecture}: Deeper networks (3-4 layers) perform best. Very deep (5 layers) shows diminishing returns. Linear model increases error by 52\%
    \item \textbf{Embedding Dimension}: 32 is optimal for this dataset. Too small (4) hurts performance (+17\% RMSE), too large (64) shows slight degradation
    \item \textbf{Regularization}: Minimal impact on this dataset. Light or no regularization performs best. KL+Dropout combination slightly degrades performance
    \item \textbf{Structure}: Hybrid structure (static SEM + dynamic attention) performs best, matching no-structure baseline while providing interpretability
    \item \textbf{Learning Rate}: 0.005 is optimal. Too low (0.0001) prevents learning, too high (0.01) causes instability. Shows 14\% improvement over default 0.001
\end{itemize}

\paragraph{Recommendations}

Based on ablation results, we recommend:
\begin{enumerate}
    \item Always include random effects for hierarchical data
    \item Use 3-4 layer networks ([128, 64, 32] for larger datasets)
    \item Set embedding dimension to 16-32 depending on group size
    \item Start with lr=0.005 and adjust if needed
    \item Use hybrid structure when interpretability is important
\end{enumerate}

\subsection{Computational Efficiency}

TabMixNN provides competitive training efficiency compared to gradient boosting methods. On the Sleepstudy dataset (180 observations, 18 groups), TabMixNN training takes approximately 2-3 seconds for 100 epochs, comparable to XGBoost training time. Inference is fast at $<$1 ms per sample. Classical \texttt{lme4} is faster for small datasets but does not scale to the nonlinear relationships that TabMixNN can capture.

\subsection{Interpretability Analysis}

Beyond predictive performance, TabMixNN provides comprehensive interpretability tools that bridge deep learning and classical statistics. We demonstrate these capabilities using the Sleepstudy dataset \citep{belenky2003patterns}, which examines reaction times of 18 subjects over 10 days of sleep deprivation with the model formula: \texttt{Reaction $\sim$ Days + (Days | Subject)}.

\subsubsection{SHAP Values for Understanding Subject-Specific Effects}

TabMixNN implements Shapley Additive Explanations (SHAP) \citep{lundberg2017unified} to quantify how features affect predictions at both population and individual levels. Rather than simply ranking feature importance, we demonstrate how SHAP reveals subject-specific heterogeneity in the effect of sleep deprivation.


\begin{figure}[htbp]
\centering
\includegraphics[width=0.95\textwidth]{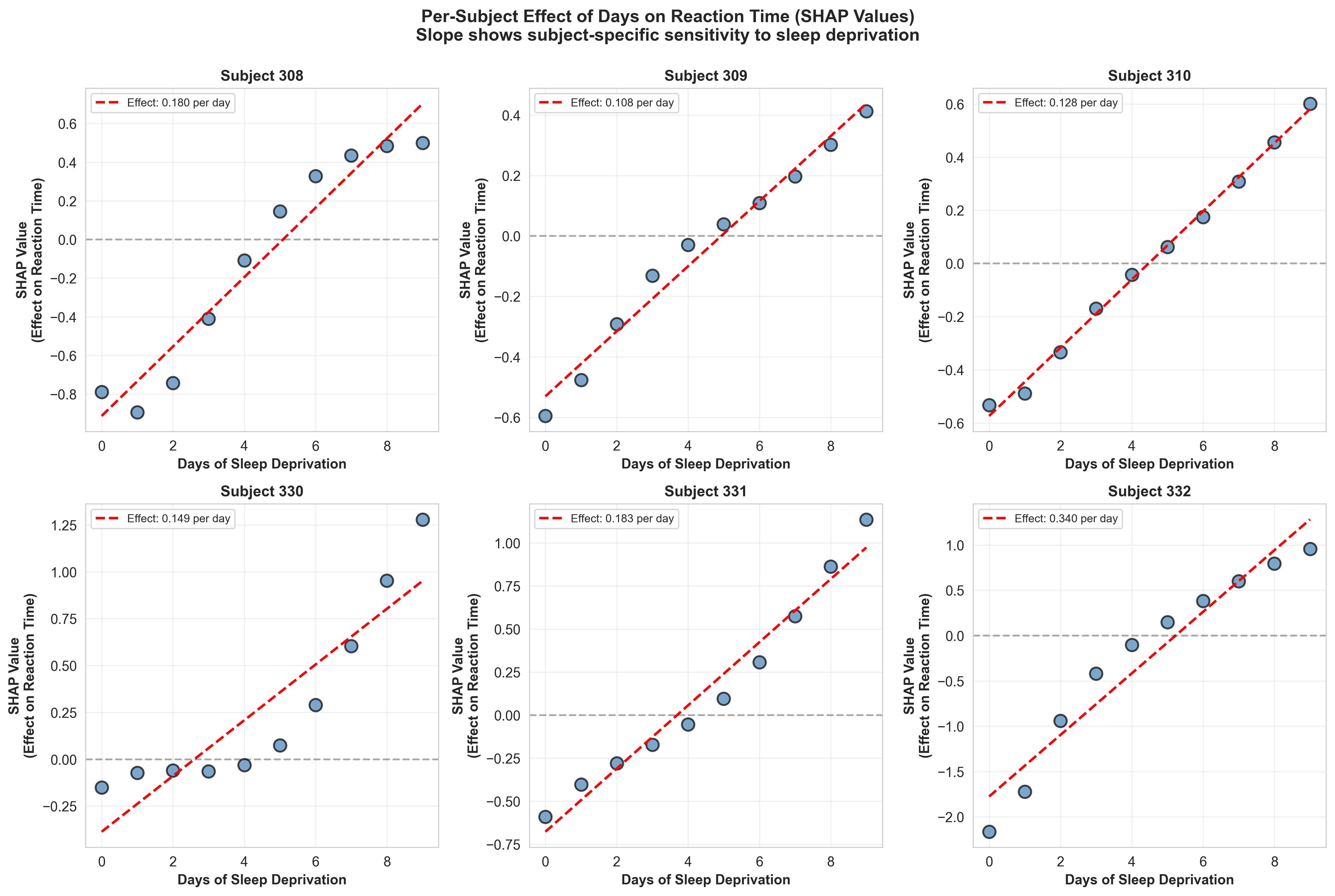}
\caption{Per-subject effect of Days (sleep deprivation) on Reaction Time using SHAP values. Each panel shows one subject's SHAP values plotted against Days, with fitted lines indicating the strength of the Days effect for that individual. Positive slopes indicate subjects whose reaction times worsen more with sleep deprivation, while negative or flat slopes indicate resilient subjects. This demonstrates heterogeneity in random slopes: different subjects show different sensitivities to sleep deprivation.}
\label{fig:shap_days_effects}
\end{figure}

\begin{figure}[htbp]
\centering
\includegraphics[width=0.95\textwidth]{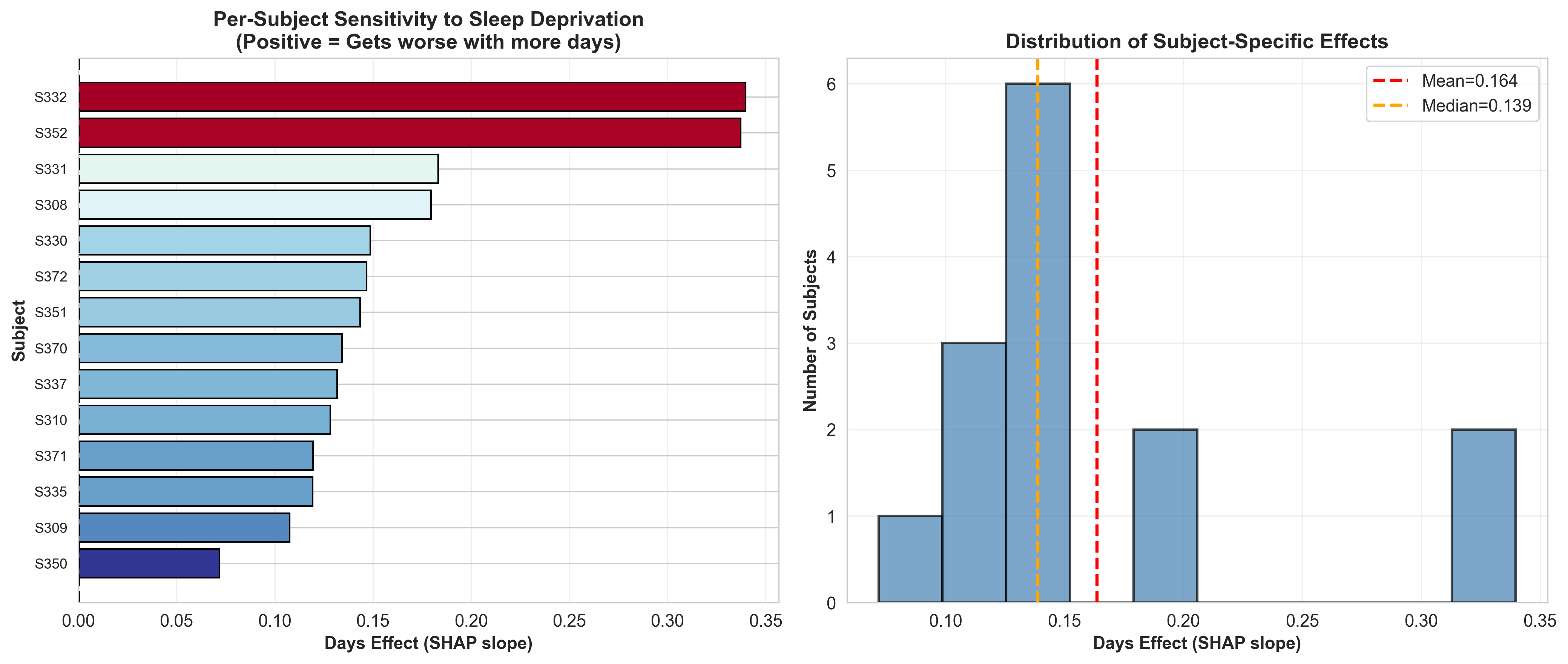}
\caption{Comparison of subject-specific Days effects across all subjects. \textbf{Left:} Bar plot showing the SHAP slope (Days effect) for each subject, sorted by magnitude. Positive values indicate subjects who get worse with more sleep deprivation. The color gradient visualizes the range of heterogeneity. \textbf{Right:} Distribution of subject-specific effects showing mean and median. This quantifies individual differences in sensitivity to sleep deprivation.}
\label{fig:shap_subject_comparison}
\end{figure}

\begin{figure}[htbp]
\centering
\includegraphics[width=0.95\textwidth]{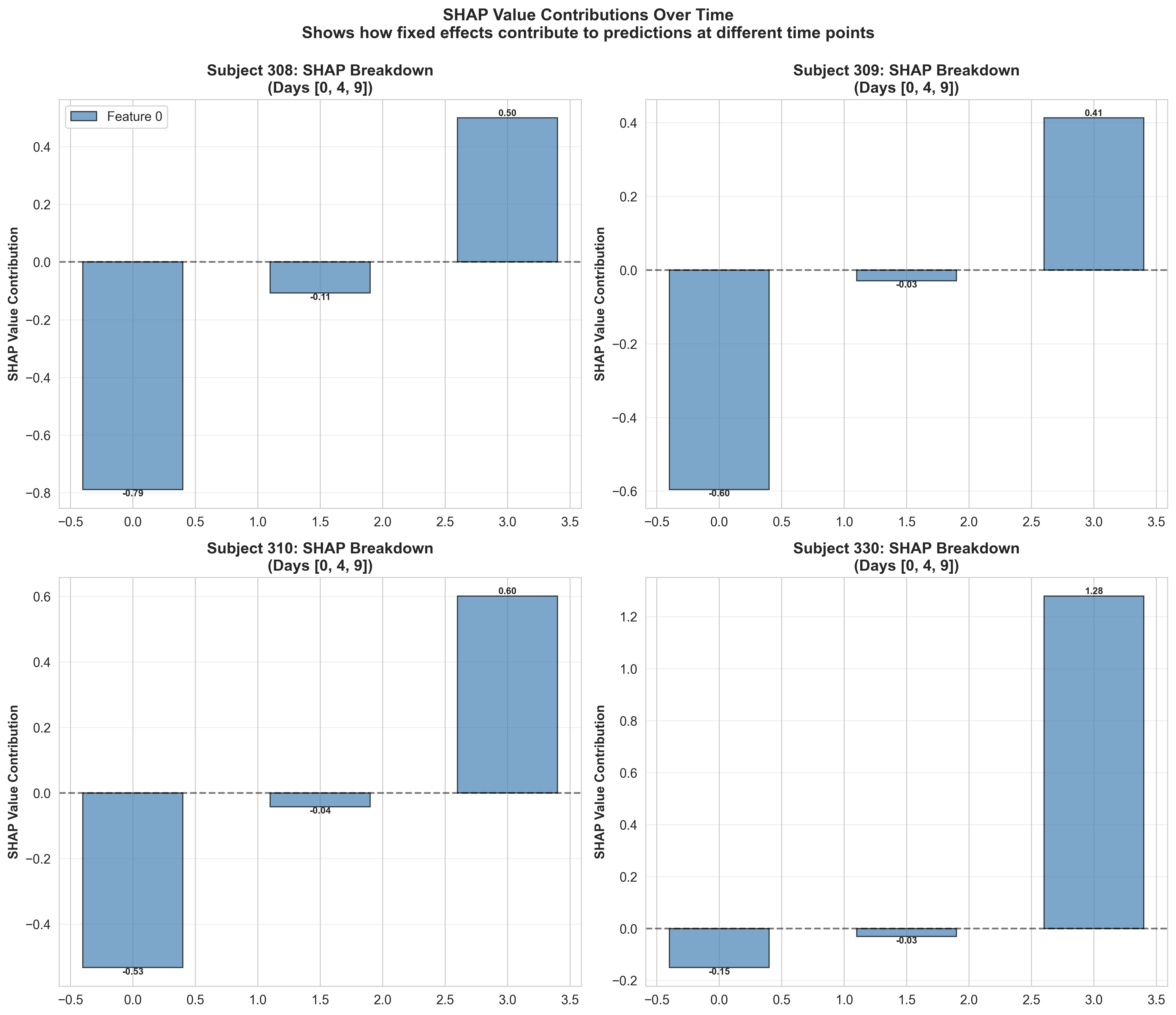}
\caption{SHAP waterfall plots showing feature contributions for four subjects at different time points (Days 0, 5, and 9). Each bar represents the SHAP contribution of a feature to the prediction at that time point. This visualization demonstrates how fixed effects (Days) and other features combine to produce predictions, and how these contributions change over time within subjects.}
\label{fig:shap_waterfall}
\end{figure}

Figure~\ref{fig:shap_days_effects} shows the \textit{effect} of Days on reaction time for six representative subjects. Each panel plots SHAP values against Days of sleep deprivation, with the fitted line's slope quantifying that subject's sensitivity. For example, some subjects show steep positive slopes (strong negative impact from sleep loss), while others show flatter or even negative slopes (relative resilience). This visualizes the random slope component of the model: $\text{Reaction}_{ij} = \beta_0 + \beta_1 \text{Days}_{ij} + (b_{0i} + b_{1i} \text{Days}_{ij}) + \epsilon_{ij}$, where $b_{1i}$ varies across subjects $i$.

Figure~\ref{fig:shap_subject_comparison} compares these subject-specific effects across all subjects. The left panel ranks subjects by their Days effect (SHAP slope), revealing substantial heterogeneity: some subjects deteriorate rapidly with sleep deprivation while others show minimal decline. The right panel shows the distribution of effects, demonstrating that individual differences are not just noise but systematic variation. This quantifies the random slope variance $\sigma^2_{b_1}$.

Figure~\ref{fig:shap_waterfall} provides a complementary view, showing how feature contributions accumulate to produce predictions for specific subjects at different time points. As Days increases from 0 to 9, the SHAP contributions grow, demonstrating the cumulative impact of sleep deprivation. The variation across subjects (different bars for each subject) again reflects individual differences in susceptibility.

This analysis demonstrates TabMixNN's unique capability among deep learning frameworks: interpretability at both the population level (Days is the main predictor) and the individual level (subjects differ in their sensitivity to Days). Unlike standard SHAP implementations that only provide feature importance, TabMixNN reveals \textit{how effects vary across individuals}, making it suitable for applications in healthcare, education, and social sciences where understanding individual heterogeneity is critical.

All interpretability analyses are available through simple API calls (\texttt{model.shapley\_values()} and \texttt{model.shapley\_random\_effects()}). Complete code examples including visualization are provided in the Supplementary Materials, Section S4.3.

\subsubsection{Structural Learning and Attention Visualization}

TabMixNN supports learning and visualizing structural dependencies through three mechanisms:

\paragraph{Static SEM Structure}

The static SEM component learns directed acyclic graph (DAG) structures among latent features via penalized optimization. The learned adjacency matrix $\mathbf{B}_s$ reveals which features directly influence others, providing interpretable causal-like relationships. Sparsity penalties encourage parsimony, typically yielding 10-30\% edge density.

\paragraph{Dynamic Attention Matrices}

The dynamic attention mechanism computes context-dependent relationships via multi-head self-attention. Attention weights reveal which features attend to which others for each prediction, capturing input-dependent dependencies. Unlike static structure, attention adapts to each sample, enabling flexible modeling of heterogeneous relationships.

\paragraph{Hybrid Structure}

The hybrid model combines static and dynamic components: $\boldsymbol{\eta} = (\mathbf{I} - \mathbf{B}_s - \mathbf{B}_d(\boldsymbol{\eta}))^{-1}\boldsymbol{\xi}$, where $\mathbf{B}_s$ captures time-invariant structure and $\mathbf{B}_d$ adds adaptive dependencies. Our ablation study (Table \ref{tab:ablation_comprehensive}) shows hybrid structure matches pure performance while providing dual interpretability: stable relationships (static) and sample-specific adjustments (dynamic).

\paragraph{Output-Level SEM for Multitask Learning}

For multitask problems, TabMixNN can learn dependencies among outcome parameters via output-level SEM. This reveals how tasks influence each other (e.g., age prediction informing maturity classification in the Abalone dataset). The learned structure $\mathbf{B}_{\text{output}}$ quantifies inter-task relationships, providing insight into task complementarity.

\section{Discussion and Conclusion}
\label{sec:discussion}

We have presented TabMixNN, a comprehensive framework for mixed-effects deep learning on tabular data. The framework successfully bridges classical statistical modeling and modern neural networks through:

\begin{itemize}
    \item \textbf{Modular architecture}: Clean separation of encoder, backbone, and output heads enables flexible composition
    \item \textbf{Rich covariance structures}: Support for temporal (AR/ARMA), spatial (GP/SPDE), and genetic (kinship) correlation
    \item \textbf{Diverse outcome families}: Unified interface for regression, classification, and multitask learning
    \item \textbf{Structural learning}: DAG constraints for causal structure discovery
    \item \textbf{Interpretability}: Parameter extraction, variance decomposition, and SHAP values
    \item \textbf{Accessibility}: R-style formula interface and scikit-learn compatibility
\end{itemize}

\subsection{Limitations and Future Work}

Current limitations and directions for future research include:

\begin{itemize}
    \item \textbf{Scalability}: For very large datasets ($n > 10^6$), additional optimizations (distributed training, approximations) may be needed
    \item \textbf{Automatic structure search}: More efficient algorithms for DAG structure learning (e.g., gradient-based search, reinforcement learning)
    \item \textbf{Uncertainty quantification}: More principled Bayesian inference (e.g., variational inference for all parameters, not just random effects)
    \item \textbf{Non-Euclidean data}: Extensions to graphs, manifolds, and other non-tabular structures
    \item \textbf{Causality}: Formal treatment of identifiability and causal inference with learned DAGs
    \item \textbf{Fairness and privacy}: Incorporation of fairness constraints and differential privacy
\end{itemize}

\subsection{Conclusion}

TabMixNN provides a flexible and interpretable framework for applying deep learning to tabular data with hierarchical structure. By combining the statistical rigor of mixed-effects models with the representational power of neural networks, it opens new avenues for modeling complex real-world data. We hope this framework will be useful for researchers across domains and facilitate the adoption of deep learning in settings where interpretability and statistical grounding are critical.

\section*{Acknowledgments}


\bibliographystyle{plainnat}
\bibliography{references}

\end{document}